\documentclass[a4paper]{styles/svproc}
\usepackage{url}

\usepackage[numbers]{natbib}

\usepackage{multicol}
\usepackage{multirow}
\usepackage[table,xcdraw,dvipsnames]{xcolor}
\usepackage[bookmarks=true]{hyperref}

\hypersetup{
    colorlinks=true,
    citecolor=Green,
    linkcolor=NavyBlue,
    filecolor=OrangeRed,      
    urlcolor=OrangeRed,
}
\usepackage{amsfonts}
\usepackage{mathtools}
\usepackage{bm}
\usepackage{xcolor}
\usepackage{booktabs}
\usepackage{subfigure}
\usepackage{enumerate}
\usepackage{fancyvrb}
\usepackage{tikz} 
\usepackage{wrapfig}
\usepackage{ifthen}

\newcommand{\net}{RuleFuser}
\newcommand\blfootnote[1]{
  \begingroup
  \renewcommand\thefootnote{}\footnotetext{#1}
  \addtocounter{footnote}{-1}
  \endgroup
}

\newcommand{\Dir}[1]{\mathrm{Dir}(#1)}
\newcommand{\prob}{\mathbb{P}}

\newcommand{\egostate}{\bm{x}}
\newcommand{\egohist}{\egostate_{t-H:t}}
\newcommand{\egofut}{\egostate_{t:t+F}}
\newcommand{\neighstate}{\bm{y}}
\newcommand{\neighhist}{\neighstate_{t-H:t}}
\newcommand{\histcontext}{\mathcal{H}}
\newcommand{\futcandidate}{\hat{\egostate}_{t:t+F}}
\newcommand{\errortrace}{\bm{e}_{t:t+F}}

\newcommand{\mapfeat}{\bm{s}}
\newcommand{\routeplan}{\bm{r}}
\newcommand{\priorclassprobs}{\bm{p}}
\newcommand{\classprobs}{\bm{q}}
\newcommand{\simplex}{\Delta^K}
\newcommand{\priorbeta}{\bm{\beta}_\mathrm{prior}}
\newcommand{\posteriorbeta}{\bm{\beta}_\mathrm{post}}
\newcommand{\evidence}{\bm{n}}

\newcommand{\latentvector}{\bm{z}}

\newcommand{\prior}{\prob(\classprobs \mid \histcontext)}
\newcommand{\posterior}{\prob(\classprobs \mid \histcontext, \evidence)}

\let\llncssubparagraph\subparagraph

\let\subparagraph\paragraph

\usepackage[compact]{titlesec}

\let\subparagraph\llncssubparagraph

\titlespacing{\section}{0mm}{1mm}{1mm}
\titlespacing{\subsection}{0mm}{1mm}{1mm}
\newcommand{\doctype}{arxiv} 

\begin{document}
\mainmatter              
\title{\net: An Evidential Bayes Approach for Rule Injection in Imitation Learned Planners and Predictors for Robustness under Distribution Shifts}


\titlerunning{\net}  
%
\author{Jay Patrikar\inst{1},
*Sushant Veer\inst{2},
*Apoorva Sharma\inst{2}, 
Marco Pavone\inst{2,3} and
Sebastian Scherer\inst{1}}
\authorrunning{Jay Patrikar et al.} 
%
\tocauthor{Ivar Ekeland, Roger Temam, Jeffrey Dean, David Grove,
Craig Chambers, Kim B. Bruce, and Elisa Bertino}
\institute{Carnegie Mellon University
\and
NVIDIA Research
\and
Stanford University
}

\maketitle              

\blfootnote{*equal advising.}

\begin{abstract}
Modern motion planners for autonomous driving frequently use imitation learning (IL) to draw from expert driving logs. Although IL benefits from its ability to glean nuanced and multi-modal human driving behaviors from large datasets, the resulting planners often struggle with out-of-distribution (OOD) scenarios and with traffic rule compliance.
On the other hand, classical rule-based planners, by design, can generate safe traffic rule compliant behaviors while being robust to OOD scenarios, but these planners fail to capture nuances in agent-to-agent interactions and human drivers' intent. \net, an evidential framework, combines IL planners with classical rule-based planners to draw on the complementary benefits of both, thereby striking a balance between imitation and safety. 
Our approach, tested on the real-world nuPlan dataset, combines the IL planner's high performance in in-distribution (ID) scenarios with the rule-based planners' enhanced safety in out-of-distribution (OOD) scenarios, achieving a 38.43\% average improvement on safety metrics over the IL planner without much detriment to imitation metrics in OOD scenarios.
\end{abstract}


\section{Introduction}

Autonomous vehicles are increasingly venturing into complex scenarios that are common in dense urban traffic. Safely navigating these scenarios while interacting with heterogeneous agents like drivers, pedestrians, cyclists, etc. requires a sophisticated understanding of traffic rules and their impact on the behavior of these agents.
However, many modern motion planners are developed by merely imitating trajectories from driving logs with no direct supervision on traffic rules. In fact, direct supervision of traffic rules is not practical as the training data would require examples of both traffic rule satisfaction and violation. Furthermore, the performance of such planners deteriorates in out-of-distribution (OOD) scenarios. On the other hand, rule-based planners account for traffic rules and are more robust to distribution shifts, but they struggle with nuanced human driving behavior (which can often violate strict traffic rules) and multi-modal intents. In this paper, we develop a neural planning framework, \net, that combines the benefits of both learning-based and rule-based planners.

\net~is inspired by the observation that, while learning-based planners can often outperform rule-based planners in in-distribution (ID) scenarios, they can behave much more erratically in out-of-distribution (OOD) scenarios.
In order to address this, \net~takes an evidential approach building on PosteriorNet \citep{charpentier2020posterior}.
Concretely, a rules-based planner (designed to comply with traffic rules) provides an informative prior over possible future trajectories an agent may take, which a learned neural network model subsequently updates to yield a data-driven posterior distribution over future trajectories. The strength of this update is controlled by an estimate of the \textit{evidence} that the training data provides for a given future being associated with the scene context. 
In this way, in OOD scenarios, the posterior will remain close to the traffic-law compliant prior, but in for ID scenes with high representation in the training dataset, the powerful learning-based model is trusted to better capture the nuances of interactive driving.
Importantly, as in PosteriorNet \cite{charpentier2020posterior}, \net~only requires access to nominal driving logs and does not require exposure to OOD scenarios.
\begin{figure*}[t]
    \centering
    \includegraphics[trim={0 1cm 0 0},clip,width=1.0\textwidth]{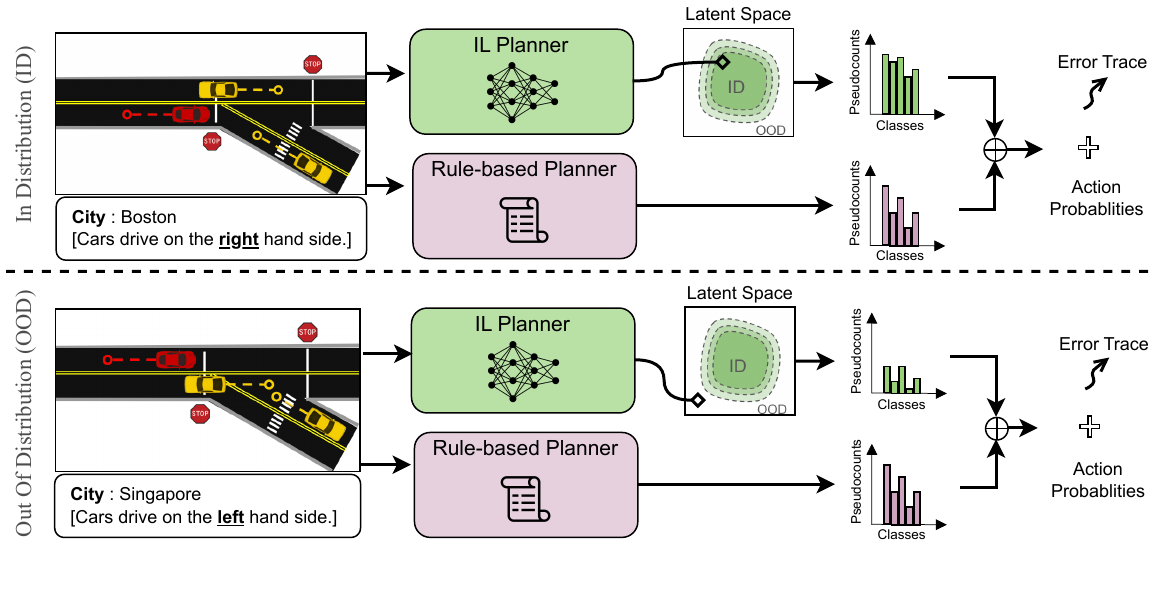}
    \caption{\net~adopts a parallel setup with two planners: a learned uncertainty-aware IL planner and a rule-based planner. 
    \textit{(Top)} For in-distribution driving data, in this illustration Boston, the IL planner learns to map the input to higher likelihood areas in latent space, indicating higher evidence. This gives the IL planner more pseudo-counts to contribute towards the Bayesian posterior contribution, surpressing the impact of the rule-based prior. 
    \textit{(Bottom)} For out-of-distribution scenarios, in this case Singapore, the input is mapped to lower evidence, so the posterior is largely controlled by the rule-based prior.}
    \label{fig:splash}
\end{figure*}
\begin{figure*}[t]
    \centering
    \includegraphics[width=0.9\textwidth]{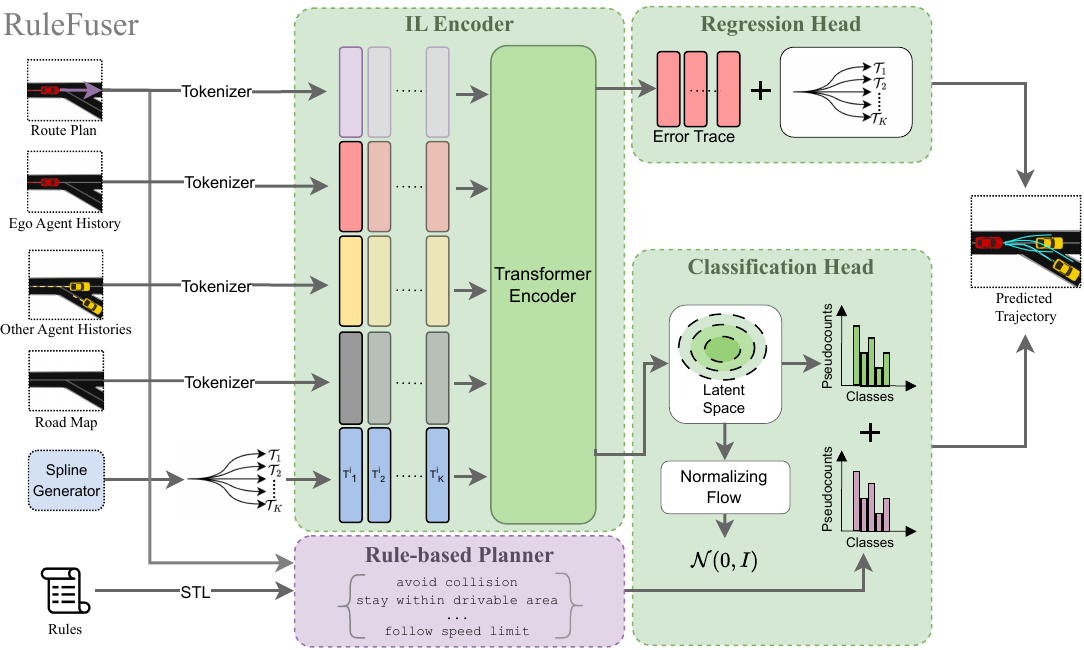}
    \caption{Overview of \net. A spline generator produces a set of dynamically generated candidate future trajectories. Each one of these trajectories is augmented with a copy of scene context inputs (ego and agent histories, map and route information), and are processed individually by a transformer-based encoder. The regression head outputs an error trace for each candidate, while the classification head estimate the the likelihood of each candidate under the training data distribution. The results are fused with prior psuedo-counts from a rule-based planner to yield the final prediction.}
    \label{fig:overview}
\end{figure*}
\begin{figure*}[t]
    \centering
    \includegraphics[width=1.0\textwidth]{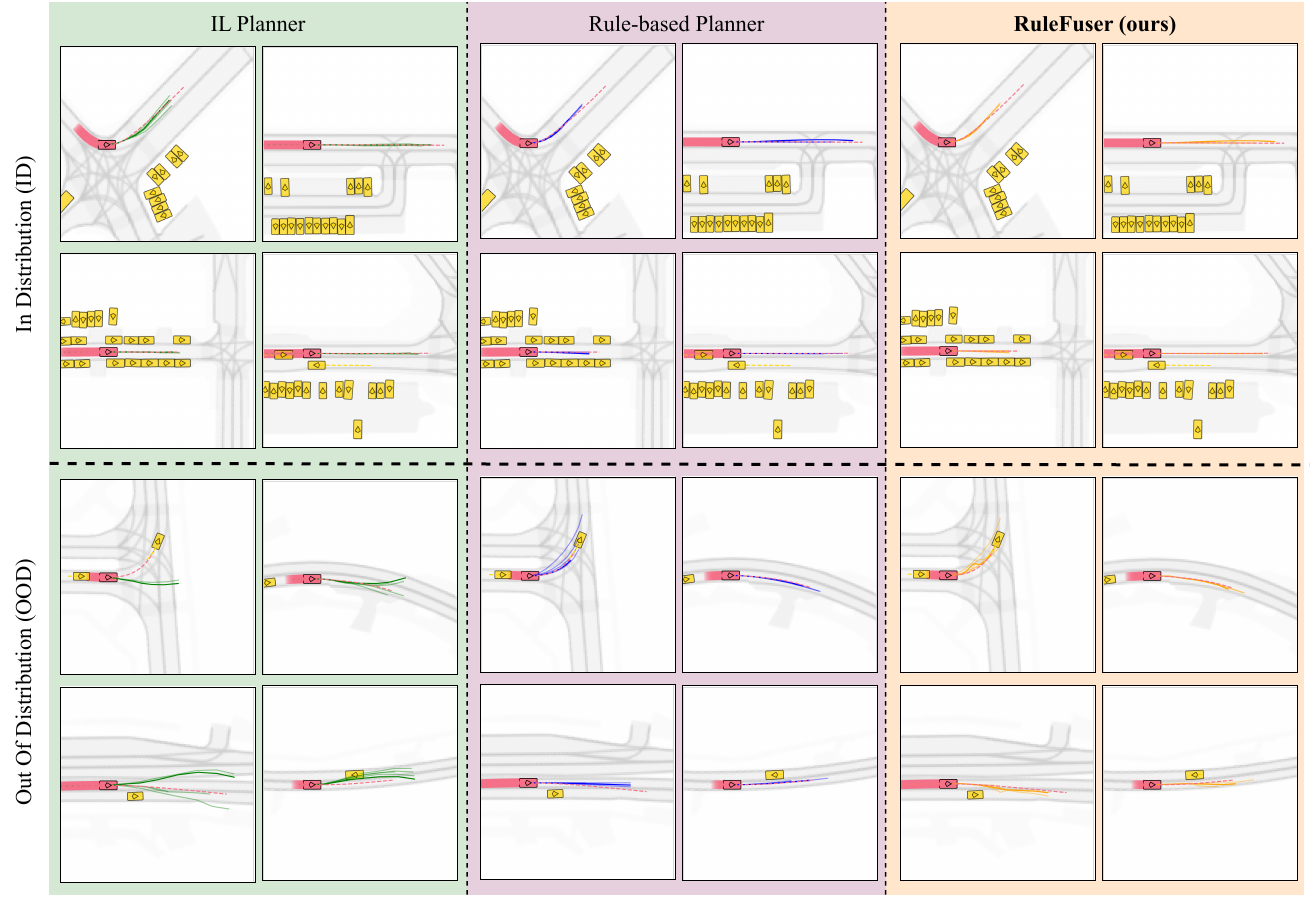}
    \caption{Figure shows the qualitative results for the three methods. While the predicted trajectories using Neural Predictor showcase good performance in Boston (ID), the performance deteriorates in the Singapore (OOD). Rule-aware Predictor has consistent performance in both ID and OOD but fails to capture nuances in speed and turning radius. \net~shows consistent performance in both ID and OOD scenarios preferring higher performance in ID and falling back to safety in OOD.}
    \label{fig:qualitative-results}
\end{figure*}

\noindent\textbf{Statement of Contributions:} The main contributions of this work are:
\begin{enumerate}
    \item We develop a novel method for injecting traffic rules in IL-based motion planners by leveraging evidential deep learning. This method permits altering the integration level between the IL and the rule-based planner without necessitating any re-training, thereby facilitating greater design flexibility.
    \item We introduce \net, a transformer-based neural framework that uses dynamic input anchor splines in a joint encoder with a posterior-net style normalizing flow decoder to estimate both epistemic and aleatoric uncertainty. We adopt the rule-based planner in \cite{veer2023multi} and use its output as a prior in a novel fusion strategy to provide robustness against OOD scenarios.
    \item We demonstrate the ability of \net~as a planner and predictor to adapt to OOD scenarios on the real-world nuPlan~\cite{nuplan} autonomous driving dataset and deliver safety levels exceeding those of learning- and rule-based frameworks alone.
\end{enumerate}




\section{Related Work}
\label{sec:related-works}

Our work builds on a rich literature on both rule-based trajectory prediction and planning, as well as uncertainty-aware learning-based trajectory forecasting and motion planning models.

\textbf{Integrating Traffic Rules in Learned Motion Planning.}
The behavior of road users is largely constrained by local traffic rules and customs. As such, it is appealing to leverage traffic law in trajectory prediction and planning. To do so requires both (i) \textit{representing} traffic law, comprised of complex temporal rules as well as a variety exceptions depending on circumstances, as well as (ii) \textit{utilizing} a representation of these rules in the motion planners. Various representations of the traffic law have been explored in the literature, such as natural language \citep{li2024driving,manas2022robust}, formal logic formulae \citep{maierhofer2020formalization,esterle2020formalizing,pigozzi2021mining}, and hierarchical rules \citep{veer2022receding,censi2019liability,tuumova2013minimum}. In this work, we lean on the hierarchical representation \cite{veer2022receding} due to its effectiveness in encoding a broader scope of traffic law while being flexible enough to permit traffic rule relaxations in the event of an exception, e.g., allowing speed limit violation to avoid collision.
Despite the interpretability of purely rule-based methods for trajectory prediction and planning, these methods can struggle in corner cases that fail to align with modeling assumptions. Conversely, while learning-based methods sidestep these modeling assumptions, their output can be erratic, and sensitive to distribution shift. Recent work proposes several strategies attempting to achieve the best of both worlds by integrating traffic rule compliance into learned motion planning. One strategy involves integrating rules into the training process, by adding a reinforcement learning (RL) objective capturing a subset of rules like collision avoidance \citep{lu2023imitation}. Similarly, \cite{li2024hydra} adds rule-supervision at train-time by training a network to predict rule-compliance of trajectory candidates in addition to predicting their likelihood, and using these predictions to re-weight candidates at inference time to favor candidates which satisfy rules over those that violate them; similarly themed approaches that explicitly use a traffic rule loss during training include \citep{zeng2020dsdnet,zeng2019end,paxton2017combining,liu2022road}. Other approaches incorporate rules only at inference-time. The \textit{safety filter} strategy, where rule-based filter directly modifies the output of a learned planner to ensure constraint satisfaction, has seen application across learning-based planning and control (see \cite{hsu2023safety} for a survey). Recently, similar strategies have been developed specifically for probabilistic planning architectures, where rule-compliance checks are used to steer the sampling process at inference time to prefer plans that meet safety constraints \citep{aloor2023follow,wang2022learning,wang2024enhancing}.
Overall, while these strategies are effective in balancing the utility of rule-based and learned motion planning, the degree to which rule-compliance shapes the output plan at test time is static. In contrast, in this work we propose a strategy which \textit{dynamically} adapts the influence of rule-compliance by estimating the epistemic uncertainty of the learned model at test time.

\textbf{Epistemic Uncertainty Quantification in Neural Networks.}
Our goal is to impose rule-based structure to learned motion planning only in those situations where we expect the learned model to perform poorly. 
Quantifying our confidence in the model is a question of estimating \textit{epistemic uncertainty}, the uncertainty in predictions stemming from limited data. There is a rich literature on quantifying epistemic uncertainty in neural network models: 
Bayesian methods, or approximations such as Monte-Carlo Dropout \cite{gal2015dropout}, Deep Ensembles \cite{lakshminarayanan2017simple}, or variational inference based approaches \cite{sharma2021sketching,Ritter2018ASL,blundell2015weight,mackay1992practical} aim to quantify epistemic uncertainty by modeling the distribution of models that are consistent with training data, but require additional computation beyond a single forward pass of the model. Other approaches aim to directly quantify epistemic uncertainty in the forward pass, either by reasoning about deviations from training data, either in the input or latent space \cite{lee_simple_nodate, zisselman_deep_2020, nalisnick_deep_2018}, directly training the model to provide an estimate \cite{van-amersfoort2020uncertainty,malinin2020regression,malinin2018predictive}, or a combination \cite{charpentier2020posterior}. 
Most related to our approach is \cite{itkina2023interpretable}, which applies the ideas in \citep{charpentier2020posterior} to the problem of  trajectory forecasting to create a model which falls back to an uninformed prediction over future agent behavior in out-of-distribution settings.
This strategy does not directly extend to motion planning, as ultimately, the vehicle must pick a single plan to execute in any given setting. 
In this work, we address this limitation by leveraging a rules-based planner as an informative prior which grounds predictions on out-of-distribution scenarios.

\section{Problem Setup and Preliminaries}
\label{sec:overview}

Let $\egostate \in\mathcal{X}$ be the state of the ego agent, $\neighstate \in\mathcal{Y}$ be the joint state for all other traffic agents in the scene, $\mapfeat \in\mathcal{S}$ be the map features (lane centerlines and road boundaries), and $\routeplan \in \mathcal{R}$ be the desired route-plan (the lanes the ego desires to track in the future). Our goal is ego motion planning: Specifically, given the past behavior of the ego $\egohist$, other agents $\neighhist$ and the map features and route plan $\mapfeat, \routeplan$, our goal is to predict the future ego behavior $\egofut$, where $H$ is the length of the past context available, and $F$ is the horizon length for prediction.
In sum, we would like a model to predict $p(\egofut \mid \histcontext)$, where we use $\histcontext := (\egohist,\neighhist, \mapfeat, \routeplan)$ as shorthand for all historical context available to the planner.
%
We structure the model's predictions by first generating a set of $K$ anchor trajectories encoding possible futures $\{\mathcal{T}_k:=\futcandidate^k \}_{k=1}^K$ by exploiting the differential flatness of the bicycle dynamics model and fitting splines connecting the current ego state to different potential future states \cite{schmerling2018multimodal}. 
Choosing these splines reduces a high dimensional regression problem to a classification problem: now, our predictions take the form of a categorical distribution $\classprobs \in \simplex$ over these anchor trajectories, where $\simplex$ represents the $K$-dimensional simplex.

We take a Bayesian perspective, and assume that we have a prior $\prior$ over $\simplex$ conditioned on the scene context $\histcontext$.
How should we use our training data to update this prior? Suppose we treat each scene context independently, and that for a particular $\neighhist$ and $\mapfeat$, we have $N$ relevant examples in our training dataset. Let $n_k$ represent the number of examples where anchor trajectory $\mathcal{T}_k$ corresponded to the true ego future, and let $\evidence$ be the vector of these counts. By Bayes rule, 
\begin{align}
\label{eq:ideal-bayes-update}
    \posterior \propto \prior \cdot \prob( \evidence \mid \classprobs, \histcontext). 
\end{align}

In this classification setting, $\prob( \evidence \mid \classprobs, \histcontext)$ is a multinomial distribution.
The conjugate prior for a multinomial distribution is a Dirichlet distribution. This means that if we parameterize the prior as $\Dir{\priorbeta}$, then the posterior according to \eqref{eq:ideal-bayes-update} will also be a Dirichlet distribution, $\Dir{\posteriorbeta}$ where $\posteriorbeta = \priorbeta + \evidence$.
We can see that the number of matching examples $N = \bm{1}^T\evidence$, also known as the total \textit{evidence}, controls the influence of the prior on the posterior prediction. Furthermore, this expression highlights why the parameter $\priorbeta$ can be interpreted as \textit{pseudo-counts} for each class.

In reality, as $\neighhist$ and $\mapfeat$ are continuous, we will not have any exact matching scenarios in our training dataset. However, scenarios are not independent from one another: training examples from similar scenes should influence our predictions on new scenes. Thus, while the Bayesian update logic above may not directly apply, we can nevertheless achieve a similar behavior by training a neural network model to estimate the evidence $\evidence$. Indeed, as in \cite{charpentier2020posterior}, we leverage a latent-space normalizing flow model to estimate the evidence $n_k$ for each anchor trajectory, thereby ensuring that in order to assign higher evidence to the regions in the latent space covered by the training data, we must take evidence away from other regions of the latent space.
In this way, in-distribution $(\histcontext,\mathcal{T}_k)$ pairs receive higher evidence, while out-of-distribution pairs are assigned lower evidence.

In this work, we propose using a rule-based planner to compute the prior pseudo-counts $\priorbeta$.
As a result, if the scene is OOD, then the total evidence $\| \evidence \|_1$ is small, and $\priorbeta$ dominates the final prediction, resulting in a smooth fallback to a rule-based planner. Conversely, if the scene is ID, then $\| \evidence \|_1$ is larger, and the final posterior relies more on the learnt network.
Fig.~\ref{fig:splash} illustrates the framework for both ID and OOD scenarios.

\newcommand{\embeddedinput}{\mathcal{C}}
\newcommand{\embeddedoutput}{\mathcal{Z}}

\section{\net}
\label{sec:rule-fuser}
In this section we introduce \net, a concrete implementation of the ideas outlined above; see Fig.~\ref{fig:overview} for an architectural overview of \net.

\subsection{Rule-Hierarchy (RH) Planner}
\label{subsec:RH-pred}
The Rule-Hierarchy planner, henceforth referred to as RH planner, is tasked with defining a prior distribution over trajectories in accordance with their compliance with a prescribed set of traffic rules. RH planner builds on the implementation in \cite{veer2023multi}, and we detail its components below:

\subsubsection{Route Planner.} The route planner generates a reference polyline by determining a sequence of lanes for the ego vehicle to follow in order to reach a desired goal location. It then combines the centerlines of these lanes into a single reference polyline.
The route planner is unaware of obstacles and other agents as well as their dynamics, it is only aware of the ego's intended position a few seconds ahead into the future and the lane graph which is made available from an HD map or an online mapping unit. The route planner chooses the nearest lane to the ego's future goal position and performs a depth-first backward tree search to the ego's current position. 

\subsubsection{Rules.} We express the traffic rules in the form of a rule hierarchy \cite{veer2022receding} expressed in signal-temporal logic (STL) \cite{maler2004monitoring} using STLCG \cite{leung2020back}. Rule hierarchies permit violation of the less important rules in favor of the more important ones if all rules cannot be simultaneously met. This flexibility allows for more human-like behaviors \cite{helou2021reasonable}. Our rule hierarchy consists of seven rules in decreasing order of importance: (i) \texttt{avoid collision}; (ii) \texttt{stay within drivable area}; (iii) \texttt{follow traffic lights}; (iv) \texttt{follow speed limit}; (v) \texttt{forward progression}; (vi) \texttt{stay near the route plan}; and (vii) \texttt{stay aligned with the route plan}.
This 7-rule hierarchy expands on the 4-rule hierarchy in \cite{veer2023multi}. Note that for evaluating the \texttt{avoid collision} rule, similar to \cite{veer2023multi}, we use a constant-velocity predictor to predict the future of non-ego agents in the scene.


\subsubsection{Trajectory Evaluation.} The rule hierarchy comes equipped with a scalar reward function $R$ which measures how well a trajectory adheres to the specifications provided in the hierarchy. The exact specifics and construction of the reward function is beyond the scope of this paper; see \cite{veer2022receding} for more details.

\subsubsection{Prior Computation.} Let $\{R_1,\cdots,R_K\}$ be the reward for the $K$ anchor trajectories. We transform these rewards into a Boltzmann distribution by treating the rewards as the negative of the Boltzmann energy. For each trajectory $i$,
\begin{align}
    [\priorclassprobs]_k = \frac{\exp(R_k/\zeta)}{\sum_{k'=1}^K\exp(R_{k'}/\zeta)} ,
\end{align}
where $\zeta$ is the Boltzmann temperature. Now, we use this distribution to define a prior by assigning pseudocounts in proportion to the probability assigned to each anchor trajectory. Let $N_\mathrm{prior}$ be a hyperparameter representing a budget on the counts. Then, we choose, $\priorbeta:= N_\mathrm{prior} \cdot \priorclassprobs$ which is then used to define the prior as $\prior := \Dir{\priorbeta}$. A large $N_\mathrm{prior}$ will require more evidence to diverge away from the prior, relying more on the RH planner, and vice-versa. Effectively, we can control how much we want to trust the rule-based prior by controlling $N_\mathrm{prior}$; we will study the effect of varying this parameter in Sec.~\ref{sec:results}.

\subsection{Evidential Neural Planner}
To complement the rule-based planner, we build an evidential learned model utilizing a Transformer-based architecture which has demonstrated strong performance in learned trajectory prediction and planning \citep{shi2022motion,ngiam2021scene,nayakanti2023wayformer}. The network takes in the scene context as well as the same $K$ anchor trajectories as the RH planner, and is tasked with estimating the \textit{evidence} the training dataset assigns to each candidate future given the scene context. The network consists of an encoder which maps the scene context and candidate future trajectory into a latent space, as well as a decoder, which uses a normalizing flow model to assign evidence to each trajectory candidate and additionally estimate an error trace to fine-tune the raw spline-based anchor trajectories. See Fig.~\ref{fig:overview} for an overview.

\subsubsection{Encoder.}
The encoder takes as input the position history of the ego $\egohist$, as well as that of nearby agents in the scene $\neighhist$. In addition, the network has access to the scene geometry $\mapfeat$ in the form of lane centerlines, road boundaries, and route information.
In addition, we also provide the network with the same $K$ anchor trajectories, $\{\mathcal{T}_k\}_{k=1}^K$, as the rule-based planner to serve as choices of ego future behavior.
First, all inputs are lifted into a $d$-dimensional space, using a learned linear map on agent trajectories, and applying PointNet \cite{qi2017pointnet} on map elements which correspond to a sequence of points (e.g., polylines).
With all inputs now taking the form of sets of $d$-dimensional vectors, we apply a sequence of Transformer blocks with factorized attention to encode these inputs. 
Specifically, we first encode the scenario history by stacking the ego history with the neighbor history, and applying attention layers which sequentially perform self-attention across time, self-attention across agents, and finally, cross-attention with the map features. As we are only interested in ego future prediction, we keep only the features corresponding to the ego history.
We apply encoding operation on the ego future candidates, and concatenate the resulting future features with the encoded ego history feature in the time dimension, yielding an embedding of shape $[K, H+F, d]$, where $H+F$ is the total number of timesteps, $K$ is the number of candidates, and $d$ is the latent feature dimension.
We apply the same sequence of factorized attention on these features to allow history features to interact with future candidate features. Note, there is no attention across different future candidates; each candidate is encoded independently.
The result is a set of features, $\embeddedoutput$, of shape $[K, H+F, d]$ 

\subsubsection{Decoder.}
The decoder is responsible for predicting the evidence, $\evidence$, to assign to each candidate as well as to predict an error trace for each anchor trajectory. 
The regression head operates directly from the temporal features, applying an MLP independently for each time-step and anchor trajectory, mapping the $d$-dimensional feature to the $2$-dimensional error vector representing the perturbation to the original spline-based trajectory.
To compute the evidence, we first average-pool the features along the temporal dimension, resulting in a $d$-dimensional feature per future candidate.
Then, an MLP projects this feature to a lower, $d_\mathrm{flow}$-dimensional space,
     $\latentvector_k = \mathrm{MLP}( \mathrm{MeanPool}( \embeddedoutput_k ) )$.
The evidence assigned to each candidate is computed using a normalizing flow with parameters $\psi$ applied over this latent space, 
    $[\evidence]_k = N \cdot p_\psi( \latentvector_k )$,
where $N$ is an evidence budget which we set to the size of the training dataset, following \cite{charpentier2020posterior}.

\subsubsection{Training Loss}
To train this model, we follow prior work on training mixture models for trajectory prediction and train for classification and regression independently.
Specifically, we impose a regression loss using a masked mean-square-error (MSE) loss function, applied only to the error trace of the mode closest to the ground truth:
\begin{align*}
    \mathcal{L}_{MSE} &= \sum_{k=1}^K \bm{1}_{k = k^*} \cdot ||(\errortrace^k + \mathcal{T}_k) -  \egofut ||^2_2.
\end{align*}
where $k^* = \arg\min_k || \mathcal{T}_k - \egofut ||^2_2$.

To train the classification head, we first convert the predicted evidence $\evidence$ into a marginal distribution over classes, $\classprobs = \evidence / \bm{1}^T \evidence$. Then, following \cite{charpentier2020posterior}, we construct a training objective by combining a classification loss on the marginal prediction $\classprobs$ with a penalty on evidence assigned to incorrect modes. Specifically, we use a binary cross entropy loss (independently over the anchor trajectories), with an entropy reward to discourage assigning evidence to incorrect classes:
\begin{align*}
    \mathcal{L}_{UCE} = \sum_{k=1}^K \mathrm{BCE}(\bar{q}_k,\bm{1}_{k=k^*}) - H(\evidence)
\end{align*}
where $\mathrm{BCE}$ is the binary cross entropy loss, and $H(\evidence)$ is the entropy of the Dirichlet distribution.

The parameters of the embedding networks, transformer encoder, decoder networks, and normalizing flow layers are all optimized through stochastic gradient descent to optimize a weighted sum of $\mathcal{L}_{MSE}$ and $\mathcal{L}_{UCE}$.

\subsection{Bayesian Fusion Strategy}
At inference time, our model fuses the outputs of the RH planner and the Evidential Neural Planner to produce an ultimate probabilistic prediction for the future agent behavior. 
First, we apply Bayes rule using our estimated evidence $\evidence$ to compute the posterior, here simplifying to $\posterior = \Dir{\posteriorbeta}$, where $\posteriorbeta = \priorbeta + \evidence$.
Incorporating the regression outputs, the posterior over the ego future behavior ultimately takes the form of a Dirichlet-Normal distribution:
\begin{align*}
    \classprobs &\sim \Dir{ \posteriorbeta }, &
    k \mid \classprobs &\sim \mathrm{Cat}(\classprobs), &
    \egofut \mid k &\sim \mathcal{N}( \mathcal{T}_k + \errortrace^k, \bm{I} ),
\end{align*}
where $\mathrm{Cat}$ indicates the categorical distribution and $\mathcal{N}(\mu, \Sigma)$ indicates the normal distribution with mean $\mu$ and covariance $\Sigma$.
Marginalizing over $\classprobs$, the final predictive distribution of our model can be viewed as a Mixture of Gaussians:
\begin{align*}
    \bar{\classprobs} &= \mathbb{E}[ \classprobs] = \posteriorbeta / ( \bm{1}^T \posteriorbeta ) \\
    p(\egofut \mid \histcontext) &= \sum_{k=1}^K \bar{p}_k \cdot \mathcal{N}( \egofut ; \mathcal{T}_k + \errortrace^k, \bm{I} ) \label{eq:likelihood}
\end{align*}
While probabilistic interpretation can be useful for trajectory prediction, when applying this model as a planner we simply return the mode with the highest posterior probability: $\mathcal{T}_{k^*} + \errortrace^{k^*}$ where $k^* = \arg\max \bar{\classprobs}$.

\section{Experiments}
\label{sec:results}

In this section, we demonstrate the ability of \net~to plan motions that balance safety and imitation in both ID and OOD data.

\subsection{Datasets and Implementation Details}
\label{subsec:datasets}
We evaluate \net~on the NuPlan \cite{nuplan} dataset, which provides labeled driving data from multiple cities around the world. We train models on data from individual cities, and subsequently test on both held-out data from the same city (ID test data), and data from another city (OOD test data). This set-up allows us to specifically test a realistic distribution shift that may arise as AVs are deployed beyond the geographical region from which training data was collected. We consider two cities for our tests, Boston and Singapore, which differ in road geometry, traffic density, and driving convention (right-side vs left-side).
For each city, \net~is trained end-to-end on the chosen train split of the NuPlan dataset. The model is trained on 8xA100 GPUs. Final network weights are chosen by their performance on the validation set from the same city. 
\subsection{Planners}
\label{subsec:predictors}
We present results for three categories of planners: IL planner, RH planner, and \net. IL planner mirrors the architecture of \net~differing only in the choice of the prior on the anchor trajectories which is always set to be uniform. Effectively, the neural model of the IL planner is ``blind" to the traffic rules and relies solely on the training data. The RH planner is the rules-based planner described in Sec.~\ref{subsec:RH-pred} that operates exclusively according to user-defined traffic rules without utilizing data. \net~integrates both approaches, as described in Sec.~\ref{sec:rule-fuser}, leveraging both the training data and user-defined rules.

\subsection{Metrics}
\label{subsec:metrics}

We use multiple metrics to compare the performance of the three planners. We separate metrics into two categories: imitation and safety.
\subsubsection{Imitation Metrics}
\begin{itemize}
    \item \textbf{ADE} (Average Displacement Error) is the average Euclidean distance between the top predicted plan and the ground truth trajectory.
    \item \textbf{FDE} (Final Displacement Error) is the smallest Euclidean distance between the terminal position of the top predicted plan and the terminal position of the ground truth trajectory.
    \item \textbf{pADE/pFDE} (Probability Weighted Average/Final Displacement Error) is the sum of the ADE/FDE over all $K$ plans, each weighted by the final categorical distribution over anchor trajectories $\bar{\classprobs}$.
    \item \textbf{Acc.} (Accuracy) is the fraction of predicted class labels that match the ground truth class label.
\end{itemize}

\subsubsection{Safety Metrics}
\begin{itemize}
    \item \textbf{Percentage Collision Rate} is the percentage of times the plan resulted in a collision with the ground truth future trajectory of another traffic agent. 
    \item \textbf{Percentage Off-Road Rate} is percentage of times the plan exited the road boundaries.
    \item \textbf{Safety Score} is an aggregate safety score based on the rank of the planned trajectory under a 2-rule rule hierarchy \cite[Definition~1]{veer2022receding} with collision avoidance and offroad avoidance as the two rules, listed in decreasing order of importance. The best rank a trajectory can get is 1 and the worst is 4. The safety score, which lies between $0$ and $100$, is computed by transforming the ranks as follows: $(100\times(\mathrm{rank}-1))/3$ where the lower safety score indicates a better safety outcome.
\end{itemize}

\subsection{Results}
\begin{figure}
    \vspace{-10pt}
    \centering
    \subfigure[Trained in Boston]{
    \includegraphics[trim=1cm 0 0 0, clip=true,width=0.45\textwidth]{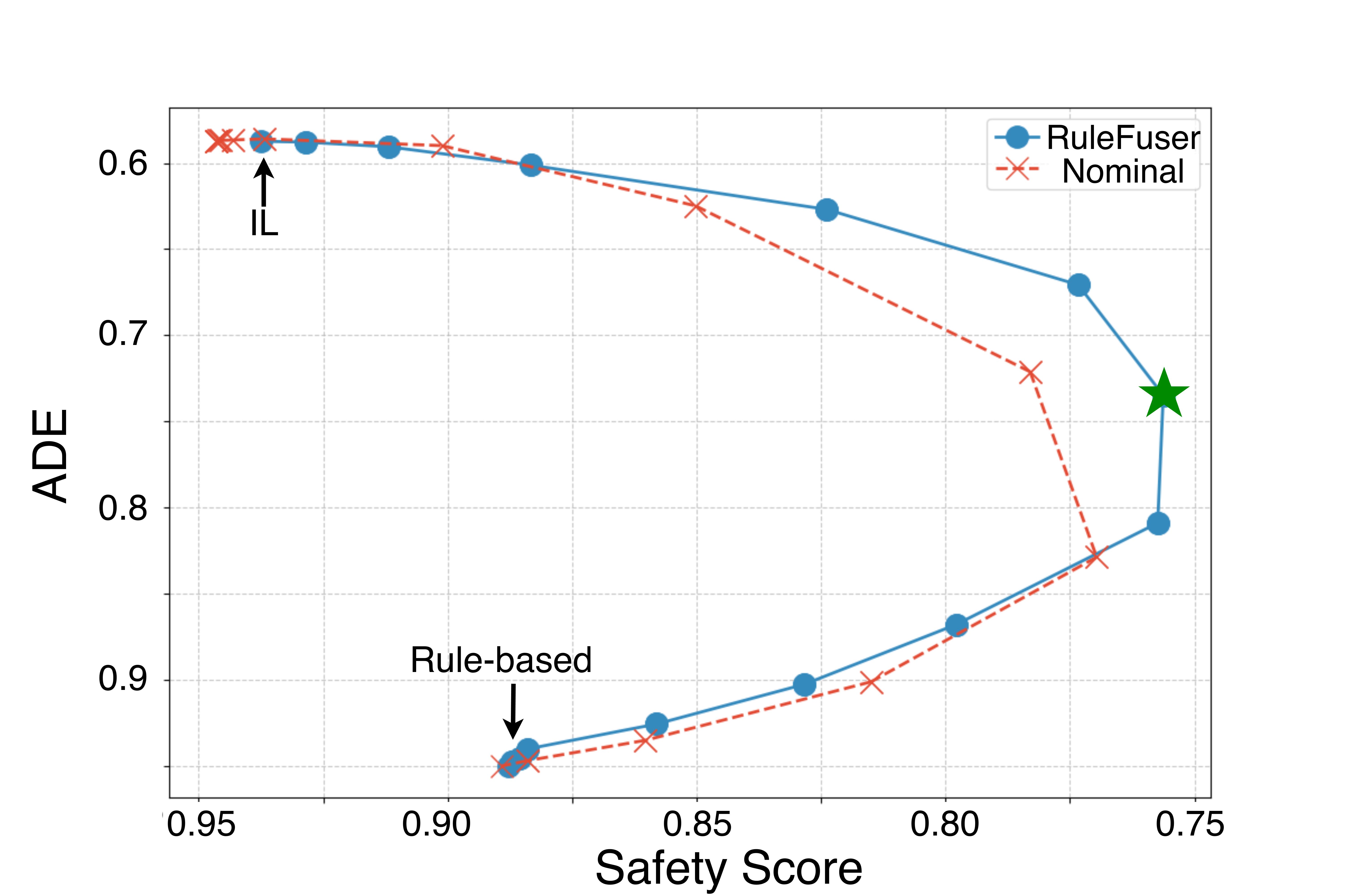}
    \label{fig:pareto-boston-train}
    }
    \subfigure[Trained in Singapore]{
    \includegraphics[width=0.45\textwidth]{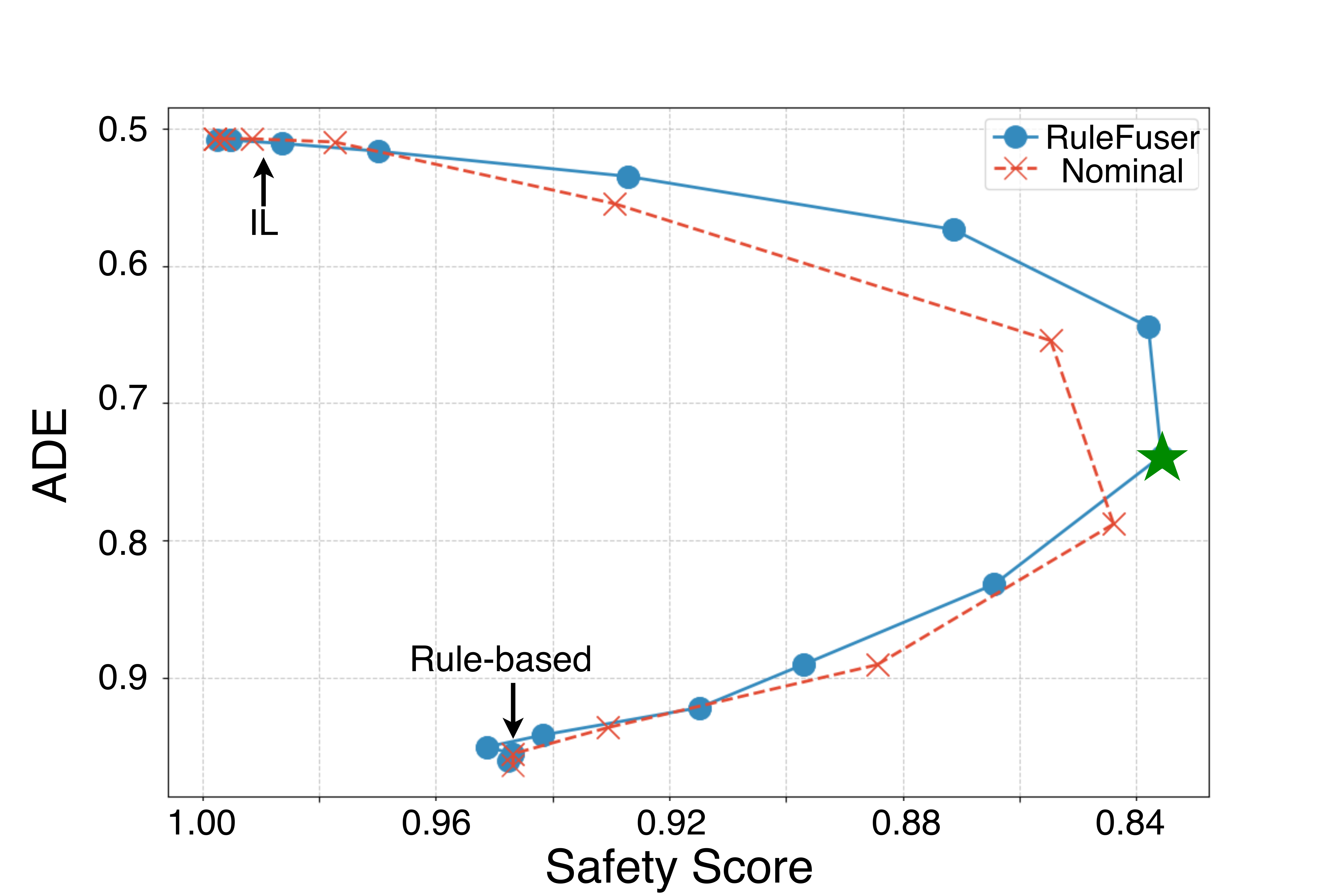}
    \label{fig:pareto-singapore-train}
    }
    \caption{\net~can generate an entire imitation-safety Pareto frontier, shown in this plot on combined ID and OOD test sets. The horizontal axis (safety score) decreases to the right, and the vertical axis (ADE) decreases upwards, so points further up and to the right are better. The bulge to the right indicates that mixing of the planners via \net~results in better safety. \net~also outperforms the nominal mixing model, indicating the importance of the evidential mixing strategy.}
\end{figure}

We present results for two different experimental setups: In one, Boston serves as the ID dataset and Singapore as the OOD dataset, while in the other, Singapore serves as the ID dataset while Boston as the OOD dataset. For both these setups, we train the IL planner on the train split of the ID dataset and evaluate the IL planner, RH planner, and 15 instantiations of \net~with different $N_{\mathrm{prior}}$ on the test splits of both the ID and OOD datasets. In Figs.~\ref{fig:pareto-boston-train} and \ref{fig:pareto-singapore-train} we plot the ADE and safety score of \net~on the combined test splits of ID and OOD datasets for each $N_{\mathrm{prior}}$. In Table~\ref{tab:qualitative} we report detailed metrics for the instantiations of \net~that exhibited the best safety score on the validation split of the respective ID datasets for both the setups; these instantiations are denoted by a star in Figs.~\ref{fig:pareto-boston-train} and \ref{fig:pareto-singapore-train}.

Additional results with prediction are available in the Appendix Table \ref{tab:pred-perf} and Table \ref{tab:pred-safety}. For the prediction task, the \net~architecture is same as planning with the only exception that route planner information is not used while training and testing the prediction model.  
\subsection{Discussion}

Our experiments provide two key findings: (i) \net~provides an effective way to balance imitation and safety; (ii) \net~exhibits safer motion plans than the IL and RH planners standalone in both ID and OOD datasets. In the rest of this section we will elaborate on these findings.

\subsubsection{\net~balances between imitation and safety.} 

\net~provides a convenient way to balance imitation and safety, allowing practitioners to control the trade-off between the two through the $N_{\mathrm{prior}}$ hyperparameter.
Figs.~\ref{fig:pareto-boston-train} and \ref{fig:pareto-singapore-train} plot an imitation-safety Pareto frontier by sweeping $N_{\mathrm{prior}}$ across a range of values. Points on the far left of the frontier are proximal to the IL planner and exhibit good imitation (lower ADE) while points on the far right are proximal to the RH planner and exhibit good safety. Notably, both the Pareto plots significantly bulge to the right indicating that mixing of the two approaches can improve overall safety beyond what a standalone planner can achieve without significantly hurting imitation. Furthermore, unlike methods which incorporate rules directly into the training objective \cite{lu2023imitation},
\net~does not require retraining to explore different points on the Pareto frontier. This makes our approach more computationally efficient for rule injection in IL planners.

To study the impact of our evidential mixing strategy for fusing IL and RH planners, we compare our approach with a nominal mixing strategy which, unlike \net, does not consider similarity of a scene to the training distribution. The nominal strategy performs a simple convex combination of the categorical distribution over candidates output by both planners,
$\classprobs := \lambda \classprobs_\mathrm{IL} + (1-\lambda)\classprobs_\mathrm{RH}$,  $\lambda\in[0,1]$.
The Pareto frontiers for the nominal mixing strategy are depicted with dashed lines in Figs.~\ref{fig:pareto-boston-train} and \ref{fig:pareto-singapore-train}. The Pareto frontier for \net~is further to the right of the nominal mixing model, validating the utility of the OOD-aware mixing strategy of \net.

\subsubsection{Impact of \net~in ID and OOD datasets.} 

\begin{wrapfigure}{r}{0.33\textwidth}
    \vspace{-20pt}
    \centering
    \includegraphics[trim=18mm 25mm 35mm 20mm, clip,width=0.33\textwidth]{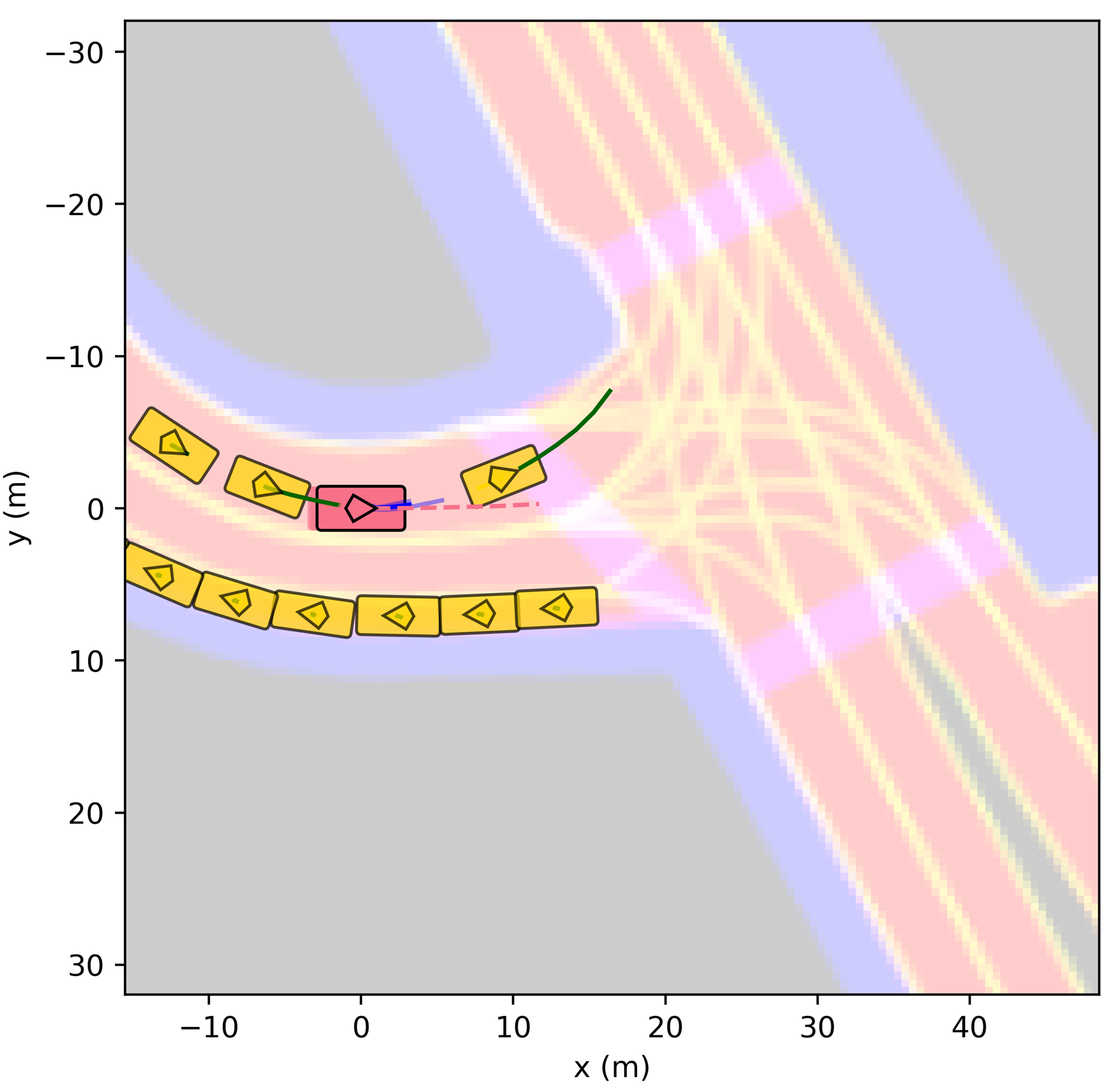}
    \caption{Rear-end collision due to constant-velocity prediction. RH Planner plans the blue trajectories for the ego vehicle (pink box). Ground truth trajectory of non-ego vehicles (yellow box) is in green.}
    \label{fig:const-vel-fail}
\end{wrapfigure}
As expected from an imitation-learned policy, the IL planner excels on imitation metrics in the ID datasets. Initially we anticipated the RH planner to outperform the IL planner on safety metrics. While the RH planner did indeed outperform the IL planner on offroad rate, it surprisingly underperformed on collision metrics; see Table~\ref{tab:qualitative}. Upon closer inspection we discovered that the underlying cause for RH planner's high collision rate was its reliance on the constant-velocity predictor which fails to account for non-ego agents' acceleration/deceleration; see Fig.~\ref{fig:const-vel-fail} for an example where ego does not plan to move forward sufficiently fast to avoid a rear-end collision due to the constant-velocity prediction for the agent behind it. An interesting insight that arises from this study is that when operating within distribution, IL approaches tend to outperform rules-based approaches when dealing with phenomenon that are challenging to model via first-principle rules, such as predicting the future trajectories of other agents. Offroad rate, on the other hand, is devoid of such modeling complexities and therefore, sees better performance from the RH Planner.

For both ID and OOD datasets, \net~always lies between the IL planner and RH planner on imitation metrics; however, \net~consistently achieves the \emph{best} safety metrics\footnote{Interestingly, in the Singapore dataset the collision rate for all planners is lower than the Boston dataset. This is the result of the much lower traffic density in Singapore with an average of 3.23 neighboring vehicles per scene as compared to Boston with an average of 11.26 neighboring vehicles per scene.}. In particular, for the OOD scenarios in Tables~\ref{tab:qualitative}(a) and (b), \net~achieves a 58.18\% and 18.67\% improvement over the IL planner, respectively (i.e., an average of 38.43\% improvement). This suggests that, in OOD scenarios, \net~effectively combines the two models to produce an overall improvement over the standalone models.
This is the outcome of \net's ``intelligent" mixing of the models which favors the model that is more suitable for a given scenario. Indeed, not every scenario in the OOD dataset will be outside the domain of competency of the IL planner; 
for example, a vehicle going straight and following a lane will behave similarly in both Boston and Singapore. 
Consequently, there are many scenarios in the OOD dataset where it is still preferable to trust the IL planner.
Indeed, as shown in Table~\ref{tab:qualitative}, \net~does not assign zero evidence in the OOD datasets. 
Remarkably, \net~in the Singapore (OOD) dataset in Table~\ref{tab:qualitative}(a) achieves a safety score of 0.23 outperforming the IL planner trained \emph{specifically} on the Singapore (ID) dataset in Table~\ref{tab:qualitative}(b) with a safety score of 0.31, providing further evidence for the ability of \net~to outperform the standalone planners on safety.

\begin{table*}
\centering
\resizebox{1.0\textwidth}{!}{
\begin{tabular}{l|ccc|ccc|c}
\multicolumn{8}{c}{\Large \textbf{(a) Boston (ID), Singapore (OOD)}} \\
\toprule
\multirow{2}{*}{\large \textbf{Metrics}} & \multicolumn{3}{c|}{\large \textbf{Imitation Metrics}} & \multicolumn{3}{c|}{\large \textbf{Safety Metrics}} & \multicolumn{1}{c}{\large \textbf{OOD}} \\
 & \multicolumn{1}{c}{\large \textbf{ADE/FDE} $\downarrow$} & \multicolumn{1}{c}{\large \textbf{pADE/pFDE} $\downarrow$} & \multicolumn{1}{c|}{\large \textbf{Acc.} $\uparrow$} & \multicolumn{1}{c}{\large \textbf{\% Coll. Rate} $\downarrow$} & \multicolumn{1}{c}{\large \textbf{\% Off. Rate} $\downarrow$} & \multicolumn{1}{c|}{\large \textbf{Saf. Score} $\downarrow$} & \multicolumn{1}{c}{\large \textbf{Evi.} $(\bm{1}^T\evidence$)} \\ 
\toprule
\large \textbf{Boston (ID):} & & & & & & & \\ 
\cellcolor[HTML]{EFEFEF}{\color[HTML]{333333} \large IL Planner} & \large \textbf{0.41}/\textbf{0.93} & \cellcolor[HTML]{EFEFEF} \large \textbf{0.46}/\textbf{1.02} & \large \textbf{0.75} & \cellcolor[HTML]{EFEFEF} \large {0.76} & \large 0.10 & \cellcolor[HTML]{EFEFEF} \large {0.55} & \large -- \\
\cellcolor[HTML]{EFEFEF}{\color[HTML]{333333} \large RH Planner} & \large 1.04/2.25 & \cellcolor[HTML]{EFEFEF} \large 1.26/2.8 & \large 0.09 & \cellcolor[HTML]{EFEFEF} \large 1.12 & \large \textbf{0.03} & \cellcolor[HTML]{EFEFEF} \large 0.76 & \large -- \\
\cellcolor[HTML]{EFEFEF}{\color[HTML]{333333} \large \net} & \large 0.70/1.55 & \cellcolor[HTML]{EFEFEF} \large 0.98/2.22 & \large 0.56 & \cellcolor[HTML]{EFEFEF} \large \textbf{0.74} & \large {0.05} & \cellcolor[HTML]{EFEFEF} \large \textbf{0.51} & \large 10.4e+6 \\ \cmidrule(lr){1-1} \cmidrule(lr){2-2} \cmidrule(lr){3-3} \cmidrule(lr){4-4} \cmidrule(lr){5-5} \cmidrule(lr){6-6} \cmidrule(lr){7-7} \cmidrule(lr){8-8}

\large \textbf{Singapore (OOD):} & & & & & & & \\ 
\cellcolor[HTML]{EFEFEF}{\color[HTML]{333333} \large IL Planner} & \large \textbf{0.76}/\textbf{1.72} & \cellcolor[HTML]{EFEFEF} \large \textbf{0.82}/\textbf{1.83} & \large \textbf{0.37} & \cellcolor[HTML]{EFEFEF} \large 0.59 & \large 0.45 & \cellcolor[HTML]{EFEFEF} \large 0.55 & \large -- \\
\cellcolor[HTML]{EFEFEF}{\color[HTML]{333333} \large RH Planner} & \large 1.08/2.32 & \cellcolor[HTML]{EFEFEF} \large 1.34/2.96 & \large 0.16 & \cellcolor[HTML]{EFEFEF} \large 0.36 & \large 0.01 & \cellcolor[HTML]{EFEFEF} \large 0.25 & \large -- \\
\cellcolor[HTML]{EFEFEF}{\color[HTML]{333333} \large \net} & \large 0.96/2.09 & \cellcolor[HTML]{EFEFEF} \large 1.18/2.64 & \large 0.30 & \cellcolor[HTML]{EFEFEF} \large \textbf{0.34} & \large \textbf{0.01} & \cellcolor[HTML]{EFEFEF} \large \textbf{0.23} & \large 3.5e+6 \\ \bottomrule
\end{tabular}
}
\vspace{5mm}
\resizebox{1.0\textwidth}{!}{
\begin{tabular}{l|ccc|ccc|c}
\multicolumn{8}{c}{} \\
\multicolumn{8}{c}{\Large \textbf{(b) Singapore (ID), Boston (OOD)}} \\
\toprule
\multirow{2}{*}{\large \textbf{Metrics}} & \multicolumn{3}{c|}{\large \textbf{Imitation Metrics}} & \multicolumn{3}{c|}{\large \textbf{Safety Metrics}} & \multicolumn{1}{c}{\large \textbf{OOD}} \\
 & \multicolumn{1}{c}{\large \textbf{ADE/FDE} $\downarrow$} & \multicolumn{1}{c}{\large \textbf{pADE/pFDE} $\downarrow$} & \multicolumn{1}{c|}{\large \textbf{Acc.} $\uparrow$} & \multicolumn{1}{c}{\large \textbf{\% Coll. Rate} $\downarrow$} & \multicolumn{1}{c}{\large \textbf{\% Off. Rate} $\downarrow$} & \multicolumn{1}{c|}{\large \textbf{Saf. Score} $\downarrow$} & \multicolumn{1}{c}{\large \textbf{Evi.} $(\bm{1}^T\evidence$)} \\ 
\toprule
\large \textbf{Singapore (ID):} & & & & & & & \\ 
\cellcolor[HTML]{EFEFEF}{\color[HTML]{333333} \large IL Planner} & \large \textbf{0.44}/\textbf{0.97} & \cellcolor[HTML]{EFEFEF} \large \textbf{0.48}/\textbf{1.08} & \large \textbf{0.67} & \cellcolor[HTML]{EFEFEF} \large {0.39} & \large 0.14 & \cellcolor[HTML]{EFEFEF} \large {0.31} & \large -- \\
\cellcolor[HTML]{EFEFEF}{\color[HTML]{333333} \large RH Planner} & \large 1.08/2.32 & \cellcolor[HTML]{EFEFEF} \large 1.34/2.96 & \large 0.16 & \cellcolor[HTML]{EFEFEF} \large 0.36 & \large 0.01 & \cellcolor[HTML]{EFEFEF} \large 0.25 & \large -- \\
\cellcolor[HTML]{EFEFEF}{\color[HTML]{333333} \large \net} & \large 0.80/1.75 & \cellcolor[HTML]{EFEFEF} \large 1.09/2.47 & \large 0.44 & \cellcolor[HTML]{EFEFEF} \large \textbf{0.29} & \large \textbf{0.01} & \cellcolor[HTML]{EFEFEF} \large \textbf{0.20} & \large 10.7e+5 \\ \cmidrule(lr){1-1} \cmidrule(lr){2-2} \cmidrule(lr){3-3} \cmidrule(lr){4-4} \cmidrule(lr){5-5} \cmidrule(lr){6-6} \cmidrule(lr){7-7} \cmidrule(lr){8-8}

\large \textbf{Boston (OOD):} & & & & & & & \\ 
\cellcolor[HTML]{EFEFEF}{\color[HTML]{333333} \large IL Planner} & \large \textbf{0.57}/\textbf{1.30} & \cellcolor[HTML]{EFEFEF} \large \textbf{0.62}/\textbf{1.43} & \large \textbf{0.45} & \cellcolor[HTML]{EFEFEF} \large 1.05 & \large 0.15 & \cellcolor[HTML]{EFEFEF} \large 0.75 & \large -- \\
\cellcolor[HTML]{EFEFEF}{\color[HTML]{333333} \large RH Planner} & \large 1.04/2.25 & \cellcolor[HTML]{EFEFEF} \large 1.26/2.8 & \large 0.09 & \cellcolor[HTML]{EFEFEF} \large 1.12 & \large \textbf{0.03} & \cellcolor[HTML]{EFEFEF} \large 0.76 & \large -- \\

\cellcolor[HTML]{EFEFEF}{\color[HTML]{333333} \large \net} & \large 0.85/1.87 & \cellcolor[HTML]{EFEFEF} \large 1.07/2.42 & \large 0.33 & \cellcolor[HTML]{EFEFEF} \large \textbf{0.90} & \large \textbf{0.03} & \cellcolor[HTML]{EFEFEF} \large \textbf{0.61} & \large 8.5e+5 \\ \hline  
\end{tabular}
}
\caption{\footnotesize Quantitative performance metrics for the IL planner, RH planner, and \net~on ID and OOD datasets. The planned trajectories have a horizon of 3 seconds. \net's performance on IL metrics always lies between the IL planner and the RH planner, but it always outperforms IL and RH planners on the safety score. Effectively, \net~provides greater safety without significant detriment to imitation.}
\label{tab:qualitative}
\end{table*}

\section{Conclusion and Future Work} 
\label{sec:conclusion}

This work introduced \net, a novel motion planning framework for balancing imitation and safety by integrating an IL planner with a rule-based (RH) planner using an evidential Bayes approach. Both planners used a common set of motion plan candidates generated on-the-fly. The RH planner provided a prior distribution over these candidates, which \net~combined with the IL planner's predictions. The fusion, guided by an OOD measure using normalizing flows, produced a posterior distribution on the motion plan candidates; the more OOD the scene, the more the posterior aligns with the prior. On real-world AV datasets, we demonstrated \net's ability to leverage the data-driven IL planner in ID scenes and the ``safer" rule-based planner in OOD scenes.


This work opens up several exciting future directions: (i) enhancing the fusion of the IL and RH planners by incorporating additional evidence beyond just the OOD measure, such as historical performance and situation criticality; (ii) conducting thorough closed-loop evaluation of \net~in nuPlan; and (iii) investigating the impact of rule injection via \net~in end-to-end planning networks like ParaDrive \cite{weng2024drive} and UniAD \cite{hu2023planning}.



\footnotesize
\section*{Acknowledgments}
This work used Bridges-2 at PSC through allocation cis220039p from the Advanced Cyberinfrastructure Coordination Ecosystem: Services \& Support (ACCESS) program which is supported by NSF grants \#2138259, \#2138286, \#2138307, \#2137603, and \#213296.

\bibliographystyle{styles/bibtex/spbasic}
\bibliography{references}
\normalsize

\ifthenelse{\equal{\doctype}{arxiv}}{
\appendix
\section{Appendix}
\subsection{Predictor Results}
We also conduct experiments to showcase the performance of \net~on trajectory prediction tasks. The prediction framework does not use the route plan while generating the outputs. In addition to reporting the performance and safety metrics already discussed in Section \ref{subsec:metrics}, we also add the following metrics:
\begin{itemize}
    \item \textbf{mADE$_k$}: (Minimum Average Displacement Error) is the smallest average Euclidean distance between the top$_k$ predicted trajectories (according to the predictor) and the ground truth trajectory.
    \item \textbf{mFDE$_k$}: (Minimum Final Displacement Error) is the smallest Euclidean distance between the terminal positions of the top$_k$ predicted trajectories (according to the predictor) and the terminal position of the ground truth trajectory.
    \item \textbf{KLDiv.}: (Kullback-Leibler divergence), or relative entropy, is the KL divergence between the predicted categorical distribution on the anchor trajectories and the one-hot ground truth categorical distribution.
    \item \textbf{NLL.}: (Negative Log Likelihood) under the marginal predictive distribution. 
\end{itemize}
\begin{table*}
\centering
\resizebox{1.0\textwidth}{!}{
\begin{tabular}{l|ccc|c|c|c|c|c}

\hline
\multicolumn{1}{l}{\textbf{Metrics [Predict]}}                    &    \multicolumn{3}{c}{\textbf{mADE/mFDE} $\downarrow$}                                                                                                                     &                    \multicolumn{1}{c}{\textbf{pADE/pFDE} $\downarrow$}                          &       \multicolumn{1}{c}{\textbf{Acc.} $\uparrow$}           &           \multicolumn{1}{c}{\textbf{KLDiv.} $\downarrow$}      &           \multicolumn{1}{c}{\textbf{NLL}$\downarrow$ }&           \multicolumn{1}{c}{ \textbf{Evidence} $(\bm{1}^T\evidence$) }                                       \\ \hline
       top$_{k}$      & \textbf{1} & \cellcolor[HTML]{EFEFEF}{\color[HTML]{000000} \textbf{5}} & \textbf{20} & \cellcolor[HTML]{EFEFEF}\textbf{All} & \textbf{All} & \cellcolor[HTML]{EFEFEF}\textbf{All} & \textbf{All} & \cellcolor[HTML]{EFEFEF}\textbf{All}  \\ \cmidrule(lr){1-1} \cmidrule(lr){2-4} \cmidrule(lr){5-5} \cmidrule(lr){6-6}   \cmidrule(lr){7-7}  \cmidrule(lr){8-8} \cmidrule(lr){9-9} 
\textbf{Boston (ID):}                         &                     &                   &                    &                                         &                    &                                            &                   &                                                         \\ 
\cellcolor[HTML]{EFEFEF}{\color[HTML]{333333} Neural Predictor} &       0.47/1.05            & \cellcolor[HTML]{EFEFEF}{\color[HTML]{000000} }    0.37/0.78               &    0.34/0.71                 & \cellcolor[HTML]{EFEFEF}    0.56/1.25           &     0.65              & \cellcolor[HTML]{EFEFEF}         4.17          &      1.16            & \cellcolor[HTML]{EFEFEF}     --                 \\
\cellcolor[HTML]{EFEFEF}{\color[HTML]{333333} RH Predictor}     &       1.30/2.84              & \cellcolor[HTML]{EFEFEF}{\color[HTML]{000000} }    0.93/1.99               &      0.52/1.08              & \cellcolor[HTML]{EFEFEF}      1.52/3.37          &     0.08              & \cellcolor[HTML]{EFEFEF}     2.41              &       1.18            & \cellcolor[HTML]{EFEFEF}   --                     \\
\cellcolor[HTML]{EFEFEF}{\color[HTML]{333333} \net}     &  0.54/1.19                   & 0.38/0.80 \cellcolor[HTML]{EFEFEF}{\color[HTML]{000000} }                   &          0.35/0.70          & \cellcolor[HTML]{EFEFEF}    0.78/1.73        &    0.64               & \cellcolor[HTML]{EFEFEF}  3.66               &           1.16        & \cellcolor[HTML]{EFEFEF}           15.1e+4               \\ \cmidrule(lr){1-1} \cmidrule(lr){2-4} \cmidrule(lr){4-4} \cmidrule(lr){5-5} \cmidrule(lr){6-6}   \cmidrule(lr){7-7} \cmidrule(lr){8-8} \cmidrule(lr){9-9}  
 
\textbf{Singapore (OOD):}                                         &                     &                                                                   &                    &                                         &                    &                                            &                   &                                                         \\ 
\cellcolor[HTML]{EFEFEF}{\color[HTML]{333333} Neural Predictor} &   0.88/1.99                  & 0.47/1.01 \cellcolor[HTML]{EFEFEF}{\color[HTML]{000000} }                   &    0.37/0.76                & \cellcolor[HTML]{EFEFEF}    0.95/2.14            &   0.30                & \cellcolor[HTML]{EFEFEF}     3.43              &       1.44            & \cellcolor[HTML]{EFEFEF}  --                   \\
\cellcolor[HTML]{EFEFEF}{\color[HTML]{333333} RH Predictor}     &     1.21/2.62                & 0.81/1.73  \cellcolor[HTML]{EFEFEF}{\color[HTML]{000000} }          &    0.51/1.06                & \cellcolor[HTML]{EFEFEF}   1.48/3.27             &               0.15     & \cellcolor[HTML]{EFEFEF}         2.57         &      1.45             & \cellcolor[HTML]{EFEFEF}    --             \\
\cellcolor[HTML]{EFEFEF}{\color[HTML]{333333} \net}     &    0.92/2.07                 & 0.46/0.99 \cellcolor[HTML]{EFEFEF}{\color[HTML]{000000} }                   &        0.37/0.76            & \cellcolor[HTML]{EFEFEF}    1.09/2.44            &                0.30    & \cellcolor[HTML]{EFEFEF}       2.99          &     1.43              & \cellcolor[HTML]{EFEFEF}  4.2e+4                             \\ \hline        
\end{tabular}
}
\caption{Table shows the quantitative performance metrics for the Neural Predictor, RH Predictor, and \net~on nuPlan's Boston (ID) and Singapore (OOD) datasets. The predicted trajectories have a horizon of 3 seconds.}
\label{tab:pred-perf}
\end{table*}

\begin{table}
\centering
\resizebox{0.75\textwidth}{!}{
\begin{tabular}{lccc|ccc}

\hline
\textbf{Metrics  [Predict]}                         &    \multicolumn{3}{c}{\textbf{\% Collision Rate} $\downarrow$}                 &       \multicolumn{3}{c}{\textbf{ \% OffRoad Rate} $\downarrow$}                                                                                        \\ \hline
       top$_{k}$      & \textbf{1} & \cellcolor[HTML]{EFEFEF}{\color[HTML]{000000} \textbf{5}} & \textbf{20} & \cellcolor[HTML]{EFEFEF}\textbf{1} & \textbf{5} & \cellcolor[HTML]{EFEFEF}\textbf{20} \\ \cmidrule(lr){1-1} \cmidrule(lr){2-4}  \cmidrule(lr){5-7} 
\textbf{Boston (ID):}                         &                     &                   &                    &                                         &                    &                                                                                                                      \\ 
\cellcolor[HTML]{EFEFEF}{\color[HTML]{333333} Neural Predictor} &       0.75            & \cellcolor[HTML]{EFEFEF}{\color[HTML]{000000} }    1.48               &     3.60               & \cellcolor[HTML]{EFEFEF}    0.15           &     0.21             & \cellcolor[HTML]{EFEFEF}         1.01                         \\
\cellcolor[HTML]{EFEFEF}{\color[HTML]{333333} RH Predictor}     &       1.16             & \cellcolor[HTML]{EFEFEF}{\color[HTML]{000000} }    1.17               &      1.90             & \cellcolor[HTML]{EFEFEF}     0.03         &     0.04              & \cellcolor[HTML]{EFEFEF}    0.21        \\
\cellcolor[HTML]{EFEFEF}{\color[HTML]{333333} \net}     &  0.66                   & 1.10 \cellcolor[HTML]{EFEFEF}{\color[HTML]{000000} }                   &          3.11          & \cellcolor[HTML]{EFEFEF}   0.15        &    0.13                & \cellcolor[HTML]{EFEFEF}  0.70           \\ \hline 
           \\ \hline 
\textbf{Singapore (OOD):}                                         &                     &                                                                   &                    &                                         &                    &                                               \\ 
\cellcolor[HTML]{EFEFEF}{\color[HTML]{333333} Neural Predictor} &   0.67                  & 0.90 \cellcolor[HTML]{EFEFEF}{\color[HTML]{000000} }                   &    1.69                & \cellcolor[HTML]{EFEFEF}    0.90           &    1.44                & \cellcolor[HTML]{EFEFEF}     3.89    \\
\cellcolor[HTML]{EFEFEF}{\color[HTML]{333333} RH Predictor}     &     0.37               & 0.43  \cellcolor[HTML]{EFEFEF}{\color[HTML]{000000} }          &    0.73               & \cellcolor[HTML]{EFEFEF}   0.01          &              0.03    & \cellcolor[HTML]{EFEFEF}          0.41             \\
\cellcolor[HTML]{EFEFEF}{\color[HTML]{333333} \net}     &    0.55               & 0.64 \cellcolor[HTML]{EFEFEF}{\color[HTML]{000000} }                   &        1.42            & \cellcolor[HTML]{EFEFEF}    0.51            &                0.75 &\cellcolor[HTML]{EFEFEF}       3.16            \\ \hline            
\end{tabular}
}
\caption{Table shows the results on quantitative safety metrics for the three predictors on both the Boston (ID) and Singapore (OOD). The predicted trajectories have a horizon of 3 seconds.}
\label{tab:pred-safety}
\end{table}

We treat Boston as ID and Singapore as the OOD dataset. The performance metrics are shown in Table \ref{tab:pred-perf} and the safe metrics are shown in Table \ref{tab:pred-safety}. Results are shown for $k = 1 ,5, 20$. 
\subsubsection{Discussion}
Inspecting Table~\ref{tab:pred-perf} shows that the Neural Predictor and \net~perform very well on prediction metrics while the RH Predictor's performance is significantly worse, which can be attributed to the RH Predictor's inability to correctly infer the intent of the agent we are predicting for. On the other hand, in Table~\ref{tab:pred-safety}, the RH Predictor generates the more rule adherent trajectories, followed by \net~and then the Neural Predictor. The key point to note here is that the prediction performance of \net~is very similar (albeit slightly degraded) to that of the Neural Predictor while its performance on the safety metrics is significantly better than the Neural Predictor. One interesting thing to note is that for ID collision rates, \net~ outperforms both the Neural Predictor as well as the RH Predictor.

As expected from a learning-based predictor, the neural predictor performs well on ID scenarios in the Boston dataset, while its performance degrades in OOD scenarios. This leads to greater volume of off-road and collision trajectories. On the other hand, RH predictor showcases consistent performance on both prediction and safety metrics in both ID and OOD scenarios. \net~demonstrates a balance between neural and RH predictors by preferring the higher performance neural predictor in ID scenarios but falling back to higher safety RH predictor in OOD scenarios. Consequently, \net's performance on the safety metrics shows a greater improvement in the OOD scenarios than in ID scenarios. 

}

\end{document}